# Using Eigencentrality to Estimate Joint, Conditional and Marginal Probabilities from Mixed-Variable Data: Method and Applications

Andrew Alojz Skabar
*Department of Computer Science and Information Technology*
*La Trobe University*
Melbourne, Australia
a.skabar@latrobe.edu.au

*Abstract*—The ability to estimate joint, conditional and marginal probability distributions over some set of variables is of great utility for many common machine learning tasks. However, estimating these distributions can be challenging, particularly in the case of data containing a mix of discrete and continuous variables. This paper presents a non-parametric method for estimating these distributions directly from a dataset. The data are first represented as a graph consisting of object nodes and attribute value nodes. Depending on the distribution to be estimated, an appropriate eigenvector equation is then constructed. This equation is then solved to find the corresponding stationary distribution of the graph, from which the required distributions can then be estimated and sampled from. The paper demonstrates how the method can be applied to many common machine learning tasks including classification, regression, missing value imputation, outlier detection, random vector generation, and clustering.

## I. Introduction

Being able to estimate joint, conditional and marginal probabilities from some dataset allows a broad range of useful tasks to be performed. For example, *classification* and *regression* involve predicting the value of some target variable conditional on the values of the other variables. Supposing that $x_1$ is the variable to be predicted, if we can estimate the conditional distribution $p(x_1|x_2, ..., x_d)$, then we can simply select the value of $x_1$ for which this distribution is a maximum [1]. However there is no reason why we would be limited to the prediction of a single variable, and we could extend this to the more general problem of *missing value imputation* [2]; for example, we could jointly impute values of two variables $x_1$ and $x_2$ by estimating the conditional distribution $p(x_1, x_2| x_3, ..., x_d)$. *Outlier detection* [3] could be performed by estimating the joint distribution $p(x_1, x_2, ..., x_d)$, and designating an observation as an outlier if its probability falls below some predetermined threshold. If we can sample values from the estimated distributions, we could perform *random vector generation* by generating full random vectors that display the same correlations as the vectors (i.e., data points) in the original data [4], [5]. If we can estimate the joint distribution for the full dataset, then we should also be able to do this for subsets of data, leading to the use of Expectation-Maximization [6] to *cluster* the data [7]. Taken together, these activities form a large chunk of the tasks commonly used in machine learning.

All of this depends, of course, on being able to estimate the various probabilities, and this is particularly challenging on datasets containing a complex mix of continuous and discrete variables. Consider, for example, the well-known Australian Credit dataset [8], which contains a total of 16 attributes, 10 of which are discrete, taking between 2 and 14 values, and 6 of which are continuous and highly skewed. Add to this the fact that the dataset is sparse, containing only 690 examples.

A general approach to estimating distributions is to specify a parameterized model for the joint distribution of variables, and to then fit the model to the data by selecting appropriate values for the model parameters. For the mixed variable case, the joint distribution is typically modelled as the product of a conditional distribution and a marginal distribution [9]. For example, one could specify the marginal distribution of the continuous variables, and multiply this by the distribution of the discrete variables conditioned on the continuous variables. Such models are known as conditional Gaussian regression models [10]. An alternative and more popular approach has been to specify the marginal distribution of the discrete variables, and to multiply this by the distribution of the continuous variables conditioned on the discrete variables. These models are known as conditional Gaussian distributions, and were originally proposed by Lauritzen and Wermuth [11]. The assumption is that there is a different multivariate normal distribution corresponding to each combination of values for the discrete variables, and this clearly leads to scalability problems in high dimensions, as the number of parameters required to model the conditional means and covariances grows exponentially with the number of variables [12]. In addition, the multinomial distribution for discrete variables will usually need to be estimated using a frequency-based approach, and if there are a large number of cells, then there may be an insufficient number of examples to estimate these frequencies reliably.

To avoid these problems, various modifications have been made to the original model proposed by Lauritzen and Wermuth [11]. For example, Edwards [13] generalized the conditional Gaussian distribution model to a hierarchical model which captures the hierarchical interactions between discrete and continuous variables, leading to a simpler parameterization for the mean and covariance. Lee & Hastie [14] and Fellinghauer et al. [15] simplified the original model by assuming that all continuous variables share a common conditional covariance. Lee & Hastie [14] further showed that under certain assumptions the full model simplifies to one involving only pairwise interactions between variables, and the model proposed by Cheng et al. [16] allows for three-way interactions between two binary and one continuous variable.

The conditional Gaussian distribution and its variants are all model-based approaches—once the parameters for the model have been determined, inferencing is performed based on these parameters, and the original data is no longer required. In contrast, memory-based—or *non-parametric*—approaches perform inferencing directly from the data. For example, the well-known *k*-nearest neighbor algorithm classifies an observation as belonging to the same class as that of the majority of its *k* nearest neighbors. There is no modelling involved—the class is determined directly from the data. In the context of probability density estimation, the analogue to the *k*-nearest neighbor classifier is to estimate the density at some point by calculating the average distance from that point to its *k* nearest neighbors—the larger the distance, the less dense the region. As is the case for the *k*-nearest neighbor classifier, if care has been taken in the scaling of variables, the selection of distance measure, and the selection of an appropriate value for the parameter *k*, then the method may provide a reasonable estimate of the density at some point.

How might the parametric and non-parametric approaches described above fare on estimating densities for the Australian Credit dataset? The full conditional Gaussian distribution model proposed by Lauritzen and Wermuth ([11], [12]) would fail dismally—not only does the estimation of the marginal distribution for the discrete variables require estimating over 40,000 probabilities from only 690 observations, but a separate set of parameters would have to be calculated for the mean and covariance of each individual Gaussian. Models such as those in [13], [14], [15] and [16] would have varying degrees of success, depending mainly on the simplifying assumptions that they make. For example, a common covariance structure may prevent a model from accurately modelling the highly skewed nature of the continuous variables. As for the non-parametric approach, the biggest challenge is likely to be in selecting a suitable distance measure. Furthermore, a measure that works well on one dataset may work poorly on another.

This paper presents an alternative non-parametric approach for estimating probabilities from mixed variable data. Rather than explicitly calculating distances between objects, the dataset is represented as a bipartite graph consisting of object nodes and attribute value nodes. An eigenvector centrality measure is then used to calculate the stationary distribution of the graph, given some initialization vector that is determined based on the distribution we wish to estimate. Probabilities are then estimated based on this stationary distribution. These probabilities can then be used to perform a wide range of machine learning tasks.

Applications of the method have been described previously. For example, [17] addressed the task of random vector generation from mixed variable data, and [18] focused on the task of clustering mixed-attribute data. While each of these tasks makes use of aspects of the method presented here, these papers were application-focused. The purpose of this paper is to place the method at center, and to present a consolidated description of how eigencentrality can be used in a flexible and intuitive way to estimate a broad range of probabilities, and to demonstrate how these probability estimates can be used within a wide range of machine learning tasks.

The paper is structured as follows. Section II describes the scheme used to represent mixed-variable datasets as graphs. Section III then provides required background on stationary distributions and eigenvector centrality. Section IV shows how conditional probability density functions (for continuous variables) and probability mass functions (for discrete variables) can be estimated from the graph, and then generalizes this to the estimation of arbitrary conditional probabilities. Section V describes the estimation of the full joint probability, and shows how this can be used to construct a likelihood function that can be maximized to determine the values of $\alpha$ and $\beta$ that parameterize the method introduced in Section IV. Section VI addresses the problem of sampling, first describing how a single random value can be sampled from the estimated distribution for that variable, and then expanding this to describe the generation of full random vectors that display the same correlations and marginal probabilities as the original dataset. Section VII then describes how the method can be applied to the tasks of classification, regression, missing value imputation, outlier detection, random vector generation, and clustering. Section VIII concludes the paper.

## II. GRAPH REPRESENTATION

### A. Notation

Notation used in the paper is provided in Table I.

TABLE I. NOTATION USED IN PAPER

| Term | Definition |
|---|---|
| $\mathbf{X}$ | Dataset |
| $N$ | Number of rows in $\mathbf{X}$ |
| $d$ | Number of columns in $\mathbf{X}$ |
| $\mathbf{x}^n$ | $n^{\text{th}}$ datapoint (row) in $\mathbf{X}$ |
| $x_m^n$ | $m^{\text{th}}$ component of datapoint $\mathbf{x}^n$ |
| $n_m$ | number of distinct values for discrete attribute $m$ |
| $a_m^j$ | $j^{\text{th}}$ value for discrete attribute $m$ |
| $\mathbf{W}$ | Graph matrix |
| $\mathbf{c}$ | Stationary distribution vector (or *centrality vector*) |
| $\boldsymbol{\theta}$ | Personalization vector |

### B. Representation of Mixed-Variable Data

It is simplest to illustrate the representation with an example. Consider a Play Tennis dataset $\mathbf{X}$ with variables (or 'attributes') *Temperature* (continuous), *Humidity* (discrete, with possible values 'high', 'normal', 'low'), *Outlook* (discrete, with values 'sunny', 'overcast', 'rainy'), and *Play* (discrete, with values 'yes', 'no'). Suppose the dataset contains two rows, corresponding to object Monday (*Temperature* = 80, *Humidity* = 'low', *Outlook* = 'sunny', *Play* = 'yes') and Tuesday (*Temperature* = 50, *Humidity* = 'high', *Outlook* = 'rainy', *Play* = 'no'). Fig. 1 shows the graph representation of this data.

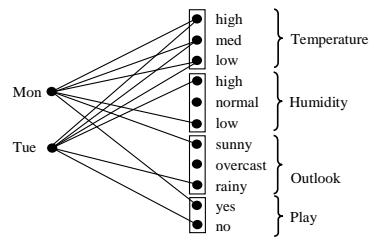

Fig. 1. Bipartite graph showing object nodes (LHS) and attribute-value nodes (RHS). An edge of weight 1 connects an object node with nodes corresponding to discrete attribute values possessed by the object. Edges connecting objects with continuous attributes (i.e., *Temperature*) use a distributed representation, where weights take some value in the interval [0, 1] (refer Fig. 2.).

The graph contains two types of nodes: *object nodes* and *attribute-value nodes*. Object nodes appear on the left-hand-side, and correspond to instances Monday and Tuesday. Attribute-value nodes appear on the right-hand side, and correspond to values that attributes may take. The graph is bipartite, meaning that edges exist only between object nodes and attribute-value nodes, but not between nodes of the same type. The graph is bi-directional, meaning that edges point from left to right, and from right to left.

For discrete attributes (i.e., *Humidity*, *Outlook* and *Play*) the representation is straightforward: an object is connected, with an edge of weight 1, to all nodes corresponding to attribute values possessed by the object.

For continuous attributes (i.e., *Temperature*) a distributed representation is used. (The rationale for using a distributed representation will be provided in Section III, in the context of describing eigenvector centrality). This distributed representation is determined using the membership functions shown in Fig. 2, where it is assumed that the values of the continuous variable have been normalized to the interval [0, 1]. Assuming that, under such a normalization, a *Temperature* value of 80 maps to 0.8, membership to the three sets is determined as High (0.6), Medium (0.4), Low (0.0). Similarly, a normalized value of 0.5 would take the distributed representation High (0.0), Medium (1.0), Low (0.0). We write these distributed representations for 0.8 and 0.5, respectively as the vectors (0.6, 0.4, 0.0) and (0.0, 1.0, 0.0). It is these distributed representation values that are used as the weighted edges linking an object to the nodes used in the distributed representation of continuous attributes.

We stress that this representation is not simply the *discretization* of a continuous variable, in which case some information would typically be lost. Under the representation described above, no information is lost, as the original continuous value can be retrieved directly from its distributed representation by simply taking the dot product of this vector with the vector (1.0, 0.5, 0.0). (i.e., 0.6 x 1.0 + 0.4 x 0.5 + 0.0 x 0.0 = 0.8).

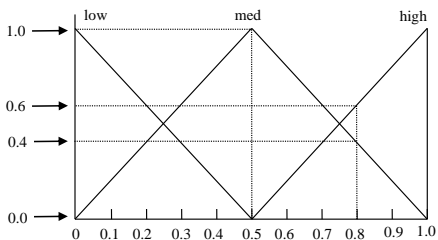

Fig. 2. Membership functions for the distributed representation of continuous variables. Assuming triples in the form [High, Medium, Low], a value of 0.8 is represented as the vector [0.6, 0.4, 0.0] and a value of 0.5 is represented as [0.0, 1.0, 0.0].

It is convenient to represent the graph as a matrix. We first construct the object-attribute table as shown in Table II and denote the matrix of values in the table as **B**.

TABLE II. OBJECT-ATTRIBUTE TABLE FOR FIG 1

|     | Temp |     |     | Humidity |     |     | Outlook |     |     | Play |     |
|-----|------|-----|-----|----------|-----|-----|---------|-----|-----|------|-----|
|     | 'h'  | 'm' | 'l' | 'h'      | 'n' | 'l' | 's'     | 'o' | 'r' | 'y'  | 'n' |
| Mon | 0.6  | 0.4 | 0.0 | 0        | 0   | 1   | 1       | 0   | 0   | 1    | 0   |
| Tue | 0.0  | 1.0 | 0.0 | 1        | 0   | 0   | 0       | 0   | 1   | 0    | 1   |

Matrix **B** has *N* rows (one corresponding to each object), and *M* columns (where *M* is the total number of attribute-value nodes required to represent the data). From **B** we then construct the adjacency matrix **W**:

$$\mathbf{W} = \begin{bmatrix} \mathbf{0} & \mathbf{B} \\ \mathbf{B}^T & \mathbf{0} \end{bmatrix}$$

Matrix **W** is a square matrix with *N+M* rows and columns. The first *N* rows and columns represent objects, and the remaining *M* rows and columns represent attribute values. The blocks of zeros in the upper left and lower right reflect the bipartite nature of the graph.

III. STATIONARY DISTRIBUTIONS AND EIGENVECTOR CENTRALITY

Having described the graph representation of mixed-variable data, we now turn attention to the estimation of various probabilities using this graph. These probabilities will be derived from the stationary distribution of the graph, which is calculated using eigenvector centrality (or *eigencentrality*), which we now discuss.

Eigenvector Centrality is based on the idea that a node's importance, or 'centrality', in a graph is high if it is connected to by other important nodes. That is,

$$c'_i = d \sum_j W_{ij} c_j \quad (1)$$

where $c_i$ is the centrality of node *i* and $W_{ij}$ is the weight of the edge connecting node *j* to node *i*. To avoid various problems that may arise in practice (see [19]), most formulations of eigenvector centrality give each node a certain amount of 'free' centrality; i.e.,

$$c'_i = d \sum_j W_{ij} c_j + (1-d) \quad (2)$$

where (1 − *d*) is the free centrality, and *d* is a value between 0 and 1. This can be conveniently expressed in vector notation as the following eigenvector equation:

$$\mathbf{c} = d\mathbf{Wc} + (1-d)\mathbf{1} \quad (3)$$

where **1** is the vector (1, 1, …).

By the Perron-Frobenius Theorem ([20], [21]), since **W** is a square matrix with real, positive entries, then the largest eigenvalue of this matrix will be real and unique. Furthermore, there will exist—up to some scaling factor—a unique eigenvector corresponding to this eigenvalue, all of whose entries are real and positive, and this will be the only eigenvector with all positive entries.

The *power iteration* method can be used to find this eigenvector: begin with a random vector $\mathbf{c}_i$, and iterate the step $\mathbf{c}_{i+1} = d\mathbf{Wc}_i + (1-d)\mathbf{1}$ until convergence, when **c** will be the dominant eigenvector, which represents the stationary distribution of the graph [19]. In this paper we refer to **c** as the *centrality vector*, and to its components as *centrality scores*.

Katz Centrality [22] and PageRank [23] are two common eigenvector centrality measures, both of which compute centrality using (3). The difference is that whereas in PageRank the matrix **W** is column normalized (i.e., normalized so that columns sum to one), in Katz Centrality no such normalization is performed. Column normalization has the effect of causing a node to distribute activation through its outgoing connections in such a way that the more outgoing connections that the node has, the less activation that will flow to the nodes it connects to; that is, its activation is shared. This is seen as beneficial in situations such as measuring importance of web pages in the World Wide Web [19], [23]. Column normalization also allows the centrality vector to be interpreted as the result of a *random walk*, since the sum of outgoing weights at some node will always sum to one. (We note that while Katz Centrality, in the strict sense, is not a random walk, it is still useful to use the idea of random walk in a looser sense due to the intuitiveness it provides).

An alternative interpretation for the second term in (3) is that it specifies the starting node for the distribution of activation throughout the graph (i.e., the start of the random walk). In (3), distribution commences equally from each node, but this need not be the case, as the vector **1** can be replaced by some vector **θ**, which can represent any arbitrary starting node or combination of starting nodes. For example, to start from some given node, we set $\theta_i = 1$ for the starting node, and 0 for all other nodes; but in general, the vector **θ** can be set to any weighted combination of nodes. The eigenvector equation now becomes

$$\mathbf{c} = d\,\mathbf{Wc} + (1-d)\,\mathbf{\theta}. \qquad (4)$$

When used in conjunction with PageRank this is often referred to as *Personalized PageRank* [24]. It is also often referred to as *Random Walk with Restart* [25]. Henceforth we will refer to the vector **θ** as the *personalization vector*.

*A. Estimating Centralities for Bipartite Graphs*

Consider the bipartite graph structure described in Section II, and suppose that we wish to calculate the centrality vector for a stationary distribution resulting from initializing the personalization vector to some object node. Activation in the graph will flow from the object node to attribute-value nodes corresponding to attribute values possessed by that object. Activation will then flow from these attribute-value nodes to other objects which possess a similar set of attribute values. We are now able to justify the distributed representation of continuous variables.

Suppose that a single weight had been used to represent continuous variables. If two objects share a high value for some continuous variable, then the activation passing through that connection will be higher than what would be the case if the objects shared the same low value for that attribute. A general treatment of continuous variables must allow for the case that the implicit similarity between two objects might be a result of their sharing *similar*, but necessarily *high*, values for some continuous attribute. To overcome this problem, we use the distributed representation scheme for continuous variables. Under this scheme, activation flow will depend only on the similarity between the values of a continuous variable, but not on the magnitude of the values.

Crucially, the eigenvector centrality measure that should be used is Katz Centrality, not PageRank. This can be understood most easily by considering the activation passing from attribute-value nodes to object nodes. The activation from some attribute-value node should be distributed equally between all object nodes possessing that attribute value (this will be the case for both Katz Centrality and PageRank); but this activation should not be less than the activation distributed by some less common attribute value. In other words, the activation from some attribute-value node must not depend on the number of outgoing connections, since we do not want to penalize attribute values that happen to be more common than others. Katz Centrality satisfies this criterion; PageRank does not.

Finally, we note that although it is not strictly necessary that the personalization vector **θ** be normalized, we will always normalize the vector so that its components sum to 1. This means that the parameter *d* that appears in (4) will always have the same effect in determining the influence of the second term relative to the first.

IV. ESTIMATING CONDITIONAL PMFS AND PDFS

This section describes the procedure for predicting the distribution of a variable conditional on the values of the other variables. There are two cases: (i) estimating the distribution of a discrete variable; i.e., estimating a *probability mass function* (pmf); and (ii) estimating the distribution of a continuous variable; i.e., estimating a *probability density function* (pdf).

*A. Discrete Variables*

Assume that $x_i$ is the discrete variable whose distribution we wish to estimate, and that the possible values that $x_i$ can take are $a_i^1, a_i^2, ..., a_i^{n_i}$. We seek, therefore, to estimate values for $p(x_i = a_i^1 | \mathbf{x}_{\setminus i})$, $p(x_i = a_i^2 | \mathbf{x}_{\setminus i})$, …, $p(x_i = a_i^{n_i} | \mathbf{x}_{\setminus i})$, where $\mathbf{x}_{\setminus i}$ denotes $x_1, ..., x_d$, but with $x_i$ omitted. These probabilities, taken together, constitute the probability mass function for $x_i$, conditioned on the values of the remaining variables. We estimate these probabilities by constructing an appropriate personalization vector **θ**, and solving (4) to find the centrality vector **c**. The procedure will be explained using the situation depicted in Fig. 3, which relates to the problem of estimating the distribution of the third variable (*Outlook*), given the values of the other three variables.

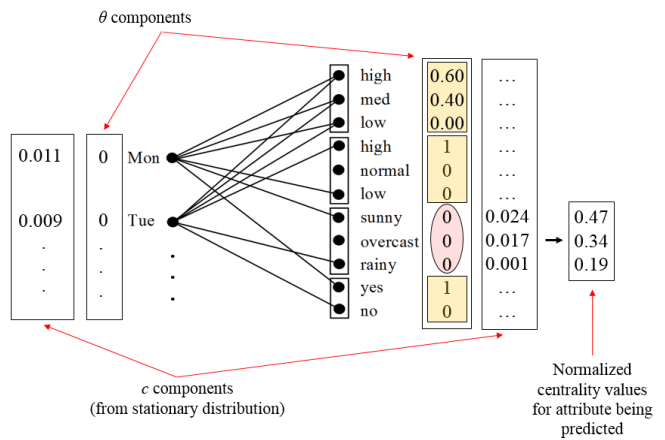

Fig. 3. Estimating the probability mass function (pmf) of discrete variable outlook (third variable), conditioned on the values of the other three variables.

The personalization vector **θ** has a component corresponding to every node in the graph. Since we wish to find the stationary distribution resulting from the attribute values that are given (i.e., the values which we are conditioning on), we set the **θ** components for object nodes to 0; we set the **θ** component of a discrete attribute value to 1 if the attribute possesses that value, and 0 otherwise; we set the **θ** components of continuous attributes to the distributed representation of that attribute; and we set the **θ** component of each of the values of the attribute whose distribution we are estimating to 0. We use a pink-shaded ellipse for the variable(s) whose distribution we wish to estimate, and yellow-shaded rectangles for the variables we are conditioning upon. Thus, in Fig. 3, we are attempting to estimate the value of the third variable, given a value of 0.8 (distributed representation [0.6, 0.4, 0.0]) for the first variable, a value of *high* for the second variable, and a value of *yes* for the fourth variable. After the eigenvector equation has been solved to find the vector $c$, we take the components of $c$ corresponding to the variable whose mass function we are estimating, and normalize these values so that they sum to one, resulting in the normalized values [0.47, 0.34, 0.19].

These normalized centrality values will be related to the probabilities that we seek, but it is useful—and indeed necessary (explained later)—to introduce a parameter to control this relationship. Let us suppose that the marginal probabilities for the values of the third attribute are [0.50, 0.30, 0.20]. These marginal probabilities can easily be determined directly from the supplied examples. We base our estimate of the conditional probabilities on the ratio of the centrality values to the marginal probabilities; i.e., on the ratio $c_i^j / m_i^j$, where $c_i^j$ is the normalized centrality of the $j^{th}$ attribute of $x_i$, and $m_i^j$ is the marginal probability of the $j^{th}$ attribute of $x_i$. If this ratio is greater (less) than 1 for some attribute value, then we expect the conditional probability for that attribute value to be greater (less) than the marginal probability. We estimate the conditional probability that attribute $x_i$ takes the $j^{th}$ value of attribute $i$ as

$$P(x_i = a_i^j \mid \mathbf{x}_{\setminus i}) = K \cdot m_i^j \left( c_i^j / m_i^j \right)^\alpha \quad (5)$$

where $K$ is a constant selected to ensure that these values sum to 1 across all attribute values; i.e.,

$$K \cdot \sum_{k=1}^{N_i} \left( m_i^k \left( c_i^k / m_i^k \right)^\alpha \right) = 1 \quad (6)$$

and the parameter $\alpha$ determines the extent to which the centrality values influence the estimate of the conditional probabilities. To illustrate, suppose that an attribute can take one of three possible values, and that the marginal probabilities and normalized centralities are as follows:

Marginal Probs ($m$):    [0.500, 0.300, 0.200]
Norm. Centralities ($c$):    [0.470, 0.340, 0.190]

Note that for the second attribute value the normalized centrality score is higher than the marginal probability, whereas for the first and third values the centrality scores are less than the marginal probabilities. The conditional distributions calculated using (5) for various values of parameter $\alpha$ are represented in Fig. 4.

When $\alpha = 0$ the estimated conditional probabilities are equal to the marginal probabilities. As $\alpha$ increases, the difference between the conditional probabilities and the marginal probabilities is accentuated, with conditional probabilities for the second value increasing, but those for the first and third value decreasing. The parameter $\alpha$ influences the 'peakedness' of the distribution.

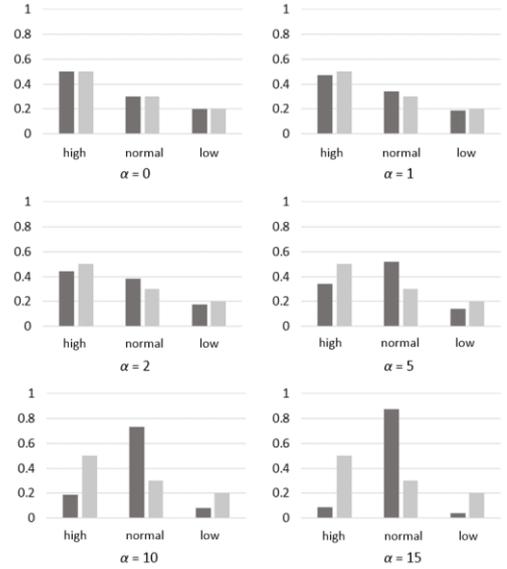

Fig. 4. Conditional (dark bars) and marginal (light bars) probabilities for variable *Outlook*. When $\alpha = 0$, conditional probabilities and marginal probabilities are equal. Increasing $\alpha$ accentuates the difference between conditional probabilities and marginal probabilities.

### B. Continuous Variables

Estimating the conditional distribution of a continuous variable is a two-stage process: we first estimate the conditional mean (i.e., mean value of the target variable conditional on the values of the other variables); we then estimate the spread about the mean.

Fig. 5 depicts the estimation of the conditional mean for variable *Temperature*. Let us assume that the variable whose density is to be estimated is $x_i$. As per the procedure for discrete variables, the personalization vector $\boldsymbol{\theta}$ is first constructed, the eigenvector equation is then solved to find the centrality vector $c$, and the components of c corresponding to variable $x_i$ are then normalized to sum to one. We emphasize that these normalized values do not represent the distribution itself; rather, they represent the distributed representation of the mean of the continuous variable, which in this case is 0.77 (i.e., 0.66 x 1.0 + 0.22 x 0.5 + 0.16 x 0.0).

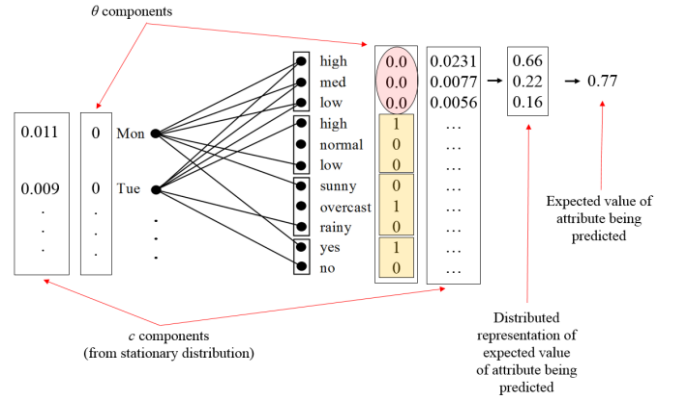

Fig. 5. Estimating the mean value of continuous variable *Temperature*, conditional on the other three variables.

For a regression problem we would simply assign $x_i$ the most likely value for the variable (0.77 in the example above). However, if we wish to sample a value for the continuous attribute, then we need to estimate an actual distribution. Since the distribution of $x_i$ is limited to an interval of finite length (recall that continuous variables are scaled to the interval [0, 1]), a suitable distribution is the Beta distribution, which is defined on the interval [0, 1], and whose shape is determined by two parameters $p$ and $q$.

$$\text{Sample } x_i \sim \beta eta(p,q) \quad (7)$$

The mean of the Beta distribution is $p/(p+q)$, and the width is controlled by $p+q$. Fig. 6 shows Beta distributions with means 0.1, 0.5 and 0.8 for three different widths. Note that the larger the value of $p + q$, the narrower is the distribution.

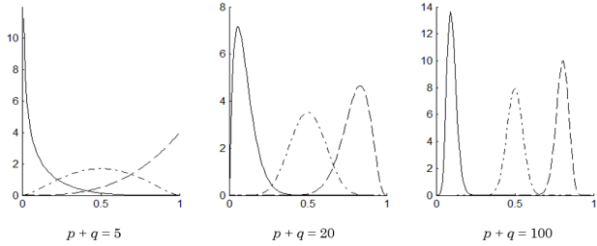

Fig. 6. Beta distributions with means 0.1, 0.5 and 0.8 for different values of parameter $\beta$ ($= p + q$).

It is convenient to introduce a parameter $\beta$ to represent the width of the distribution (i.e., $\beta = p + q$). Note that distribution parameters $p$ and $q$ can be determined since the mean of the distribution $p/(p+q)$ will equal the conditional mean $E(x_i)$, and the value of $\beta$ will be specified. The parameter $\beta$ can be considered analogous to the parameter $\alpha$ used for discrete attributes.

### C. Other Conditional Distributions

The previous two subsections addressed the problem of estimating the distribution of some variable $x_i$ conditional on *all* the remaining variables; however, the method can also be used to estimate *any* conditional distribution, and is simply a matter of constructing the appropriate personalization vector **θ**, all other details remaining the same. For example, Fig. 7 depicts the situation for estimating the distribution of the fourth variable, conditional only on the value of the second variable being 'normal' and third variable being 'sunny'. Distributions conditional on two, three, or any other number of variables can be estimated in exactly the same way.

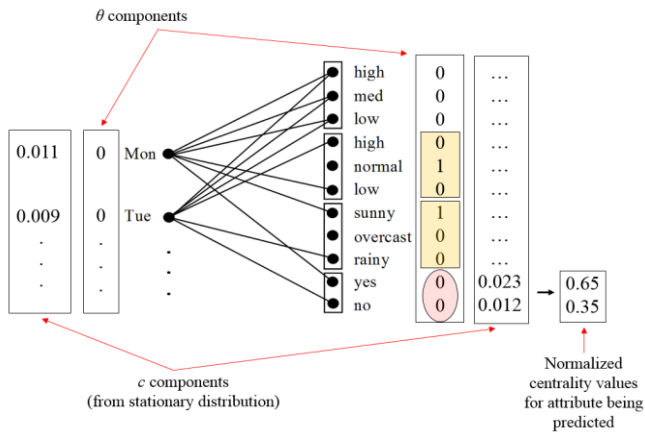

Fig. 7. Estimating the probability mass function (pmf) of discrete variables *Play*, conditional on the values of *Outlook* and *Humidity*.

## V. THE JOINT PROBABILITY DISTRIBUTION

The joint probability $p(\mathbf{x}) = p(x_1, x_2, ..., x_d)$ can be expressed as the probability chain

$$\begin{aligned} p(\mathbf{x}) &= p(x_1 | x_2, ..., x_d) \\ &\times p(x_2 | x_3, ..., x_d) \\ &\vdots \\ &\times p(x_{d-1} | x_d) \\ &\times p(x_d). \end{aligned} \quad (8)$$

The probability $p(x_d)$ is the marginal probability for $x_d$ and can be estimated directly from the data; all other probabilities are conditional probabilities, and can be estimated using the methods described in Section IV.

Being able to estimate the joint distribution is useful because it provides a basis for tasks such as outlier detection, where the objective is to identify data points that are unlikely to have occurred (see Section VII.C). It also allows the maximum–likelihood optimization of parameters $\alpha$ and $\beta$, which control, respectively, the peakedness of the conditional pmf for discrete variables and the conditional pdf for continuous variables.

### A. Maximum Likelihood Estimation of α and β

We wish to find the most likely values of $\alpha$ and $\beta$, given data **X**. This is equivalent to finding the values of $\alpha$ and $\beta$ which maximize the probability of observing **X**; that is,

$$\mathcal{L}(\alpha, \beta | \mathbf{X}) = p(\mathbf{X} | \alpha, \beta) \quad (9)$$

**X** consists of $N$ vectors $\mathbf{x}^1$, $\mathbf{x}^2$, …, $\mathbf{x}^N$. If these vectors are drawn independently from the distribution $p(\mathbf{x} | \alpha, \beta)$, then the likelihood can be expressed as

$$\mathcal{L}(\alpha, \beta | \mathbf{X}) = \prod_{n=1}^{N} p(\mathbf{x}^n | \alpha, \beta) \quad (10)$$

and it is this value which we must maximize. Equivalently, we can maximize the log likelihood:

$$\ln \mathcal{L}(\alpha, \beta | \mathbf{X}) = \sum_{n=1}^{N} \ln p(\mathbf{x}^n | \alpha, \beta) \quad (11)$$

We demonstrate the optimization of $\alpha$ and $\beta$ in Section VII.A.

## VI. SAMPLING FROM ESTIMATED DISTRIBUTIONS

There are situations where instead of choosing the mean value of some variable, we may wish to *sample* a random value from the distribution.

### A. Sampling (Single) Values

Sampling a single value for a discrete variable is straightforward. First estimate the conditional pmf for the target variable. Then construct the cumulative pdf. Finally, generate a random number from a uniform distribution on [0, 1], and use this to select a value from the cumulative pmf. The result will be a random value sampled from the pdf. For continuous variables, use the method of Section IV.B to estimate the conditional mean of the target variable, and then sample a value from the Beta distribution with this value as its mean.

## B. Sampling (Complete) Vectors

Full random vectors which display the same correlations and marginal distributions as examples in the dataset can be generated as follows. First sample the value of one of the components—say, the last, $x_d$—from the marginal distribution of that variable, $p(x_d)$. Then sample a value for the second last variable, $x_{d-1}$, conditional on the value of all the variables that have been sampled thus far. Continue this procedure until an entire vector has been generated. The process can be depicted as follows:

$$\text{Sample } x'_d \sim p(x_d)$$
$$\text{Sample } x'_{d-1} \sim p(x_{d-1}|x'_d)$$
$$\vdots$$
$$\text{Sample } x'_2 \sim p(x_2|x'_3,...,x'_d)$$
$$\text{Sample } x'_1 \sim p(x_1|x'_2,...,x'_d)$$

where $(x'_1, x'_2, ..., x'_d)$ is the resulting random vector.

## C. Sampling from Arbitrary Marginal or Conditional Distributions

The method can also be used to generate vectors from any marginal or conditional distribution. To generate vectors from a marginal distribution, simply generate vectors as described above (i.e., over the full joint distribution), but keep only the variables of interest. To generate vectors from conditional distributions, clamp the variables to be conditioned on, sampling only the remaining variables.

## VII. APPLICATIONS

In this section we describe how the methods described above can be applied to a broad range of common tasks. Where appropriate, we demonstrate using the Australian Credit dataset. As previously mentioned, this dataset contains 690 examples, described over 16 attributes (10 discrete and six continuous). To illustrate the complexity of the dataset, marginal distributions for the 16 variables are shown in Fig. 12. The dataset has traditionally been used as a classification dataset, where the objective is to predict the value of (binary) Variable 16.

### A. Maximum Likelihood Optimization of $\alpha$ and $\beta$

As described in Section IV, there are two parameters that need to be determined: $\alpha$, which controls the peakedness of the estimated probability mass functions for discrete variables, and $\beta$, which controls the width of estimated density functions for continuous variables. Fig. 8 shows the log-likelihood values corresponding to various combinations of values $\alpha$ and $\beta$ for the Australian Credit dataset. The maximum likelihood occurs for values of approximately 9 and 6 respectively for $\alpha$ and $\beta$.

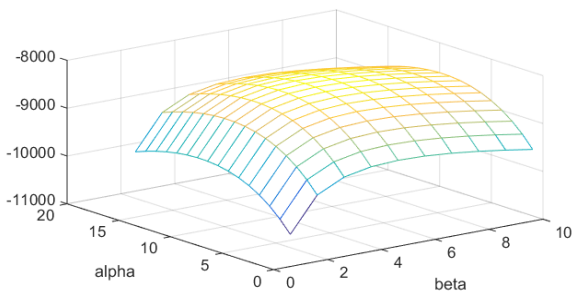

Fig. 8. Log-likelihood $\ln \mathcal{L}(\alpha,\beta|\mathbf{X})$ for the Australian Credit dataset.

### B. Classification and Regression

Classification and regression are the simplest applications of the method as they involve predicting the most likely value of a single variable given the values for the remaining variables. Thus, assuming we wish to predict the value for variable $x_1$, we need simply estimate the conditional distribution $p(x_1|x_2, x_3, ..., x_d)$. This will be either a probability mass function (pmf) or probability density function (pdf), depending on whether variable $x_1$ is discrete or continuous respectively. For a classification problem we select the most likely value from the pmf; for a regression problem we use the conditional mean of the pdf. Estimating the distribution will require solving an eigenvector equation once for each value to be predicted.

We have applied the method to the Australian Credit dataset. Using leave-one-out cross-validation results in a classification accuracy of 85.7%. The accuracy achieved with a logistic regression classifier was 85.5%.

### C. Outlier Detection

Outliers are observations that are distant from other observations. In a probabilistic context they are points for which the probability of occurrence is very small.

The probability with which some observation occurs is given by the value of the joint probability at that point. Given some point $\mathbf{x} = (x_1, x_2, ..., x_d)$, we estimate the probability of observing this point, $p(\mathbf{x})$, and designate the point as an outlier if the probability falls below some selected threshold. The method described in Section V can be used to estimate the joint probability.

Fig. 9 contains a histogram showing distribution of the log of joint probabilities for the Australian Credit dataset. There are a small number of examples for which the joint probability is very small. These might be considered outliers.

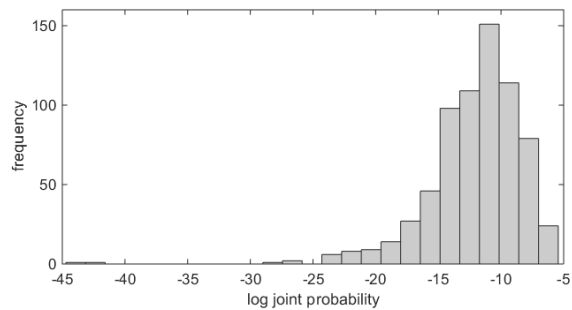

Fig. 9. Histogram showing the distribution of the log of joint probabilities of the 690 examples in the Australian Credit dataset. Examples with a very low probability can be considered outliers.

### D. Missing Value Imputation

Classification and regression can be considered special cases of the missing value imputation problem where the objective is to assign the most likely value of a *single* variable. More generally, however, missing value imputation may involve assigning a value for two or more variables.

Imputing the value of two or more missing values is straightforward. Suppose, for example, that a data point is missing values for variables $x_1$ and $x_2$, but all other values are known. There are two options. We could impute the most likely values for $x_1$ and $x_2$ from the distributions $p(x_1|x_3,...,x_d)$ and $p(x_2|x_3,...,x_d)$ respectively.

Alternatively, we could impute random values. This can be done by sampling a value for one of the variables, say $x_2$, from the distribution $p(x_2|x_3,…, x_d)$, and then sampling a value for $x_1$ from the distribution conditional on this value of $x_2$; i.e., the distribution $p(x_1|x_2,…, x_d)$. Clearly this procedure can be generalized to the generation of any number of missing values. In the limiting case where *all* values are missing, the problem becomes that of *random vector generation*, which we now discuss.

*E. Random Vector Generation*

Given a dataset **X**, random vector generation is the task of generating random data points (vectors) distributed just like the vectors in **X**. The generated vectors should have the same marginal distributions as those in **X**, and should display the same correlations as the data in **X**. There are many uses for random vector generation, the most obvious being for generating data for simulation studies.

Random vectors can be generated using the method described in Section VI.*B*. However, if we are required to generate a large number of vectors, then it may be more efficient to use Gibbs Sampling [26]. Suppose that $p(\mathbf{x}) = p(x_1,...,x_d)$ is a distribution from which we wish to sample. At each step of the Gibbs sampling procedure, the value of one of the variables, say $x_i$, is replaced by a new value, drawn from the distribution of $x_i$ conditioned on the values of the remaining variables; that is, $x_i$ is replaced by a value drawn from $p(x_i | \mathbf{x}_{\setminus i})$. This procedure is then repeated by cycling through the remaining variables.

**ALGORITHM 1.** Gibbs Sampling Procedure
Initialize $\{x_i : i = 1, …, D\}$
for $\tau = 1,...,T$
  Sample $x_1^{(\tau+1)} \sim p(x_1 | x_2^{(\tau)}, x_3^{(\tau)},...,x_D^{(\tau)})$.
  Sample $x_2^{(\tau+1)} \sim p(x_2 | x_1^{(\tau+1)}, x_3^{(\tau)},...,x_D^{(\tau)})$.
  ⋮
  Sample $x_j^{(\tau+1)} \sim p(x_j | x_1^{(\tau+1)},...,x_{j-1}^{(\tau+1)}, x_{j+1}^{(\tau)},...,x_D^{(\tau)})$.
  Sample $x_D^{(\tau+1)} \sim p(x_D | x_1^{(\tau+1)}, x_2^{(\tau+1)},...,x_{D-1}^{(\tau+1)})$.
end for

The entire procedure is then repeated continually until a sufficient number of examples have been drawn.

*1) Play Tennis Dataset*

As a demonstration, we apply random vector generation to a modified version of the Play Tennis dataset in which variables *Temperature* and *Humidity* are both continuous, and variable *Outlook* is discrete. The demonstration will also serve to present a visualization of the effect of parameters $\alpha$ and $\beta$ on the estimated conditional distributions.

Fig. 10 shows a plot of 10 supplied data points. Attributes *Outlook*, *Temperature* and *Humidity* are each represented on a separate axis. The left plot shows examples for which the value of *Play* is 'no' and the right shows points for 'yes'.

The distribution of the randomly generated vectors will depend on the values of $\alpha$ and $\beta$ used in the estimation of the conditional distributions. Figs. 11(a) to 11(c) each show plots of 500 random vectors, generated using different values for parameters $\alpha$ and $\beta$.

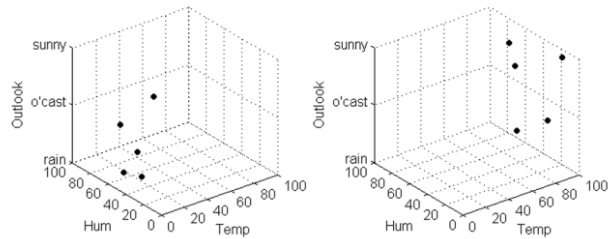

Fig. 10. Data points for the Play Tennis dataset. Left plot corresponds to *Play Tennis* = No; right corresponds to *Play Tennis* = Yes.

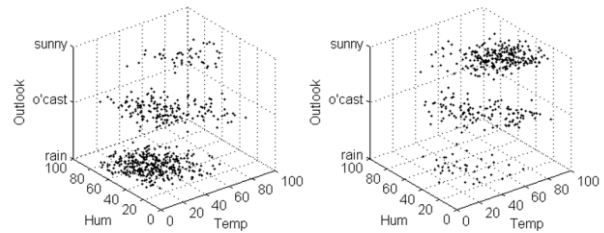

(a) Generated data ($\alpha$ = 3, β =12)

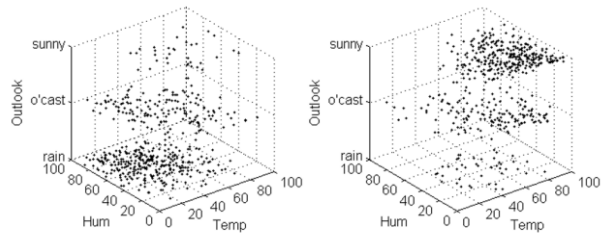

(b) Generated data ($\alpha$ = 3, β = 6)

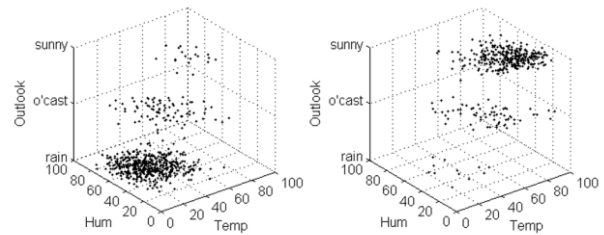

(c) Generated data ($\alpha$ = 5, β =12)

Fig. 11. Randomly generated points for the Play Tennis dataset.

The plots clearly show that the distributions responsible for the randomly generated points in Figs. 11(a) to 11(c) are quite feasibly the distributions that might be responsible for generating the 10 supplied examples in Fig. 10. The plots also demonstrate the effects of parameters α and β. The values of the continuous attributes in Fig. 11(b) (β = 6) have a greater spread than those in 11(a) (β = 12), and the values for discrete attribute Outlook have a greater spread in Fig. 11(a) (α = 3) than those in 11(c) (α = 5).

*2) Australian Credit Dataset*

We have also applied the method to generating random vectors for the Australian Credit dataset. While it is relatively straightforward to verify that the generated examples display the same marginal probabilities as the examples in the original dataset, it is not straightforward to compare the correlations between variables in the generated examples with the correlation in the original dataset, since this would involve comparing some 120 correlations at just the pairwise level. Moreover, this would involve measuring correlations between discrete and continuous variables, which may be difficult.

As an alternative, we have trained a logistic classifier using the randomly generated examples as a training set, and the original dataset as the test set. If the classifier performs successfully in classifying the examples in the original dataset, then we can conclude that the generated examples reflect the correlation structure of the original dataset.

The random vector generation method was used to generate 5000 random vectors. These randomly generated examples were then used to train a logistic regression classifier. When applied to the test examples (i.e., the 690 examples in the original dataset), a classification accuracy of 85.4% was obtained, which is virtually identical to the training accuracy obtained using logistic regression on the original examples (see Section VII.*B*), thus confirming that the two sets of examples display the same correlation structure. Fig. 12 shows that marginal distributions of variables in the randomly generated vectors are the same as those in the original vectors. Additional results and discussion can be found in [17].

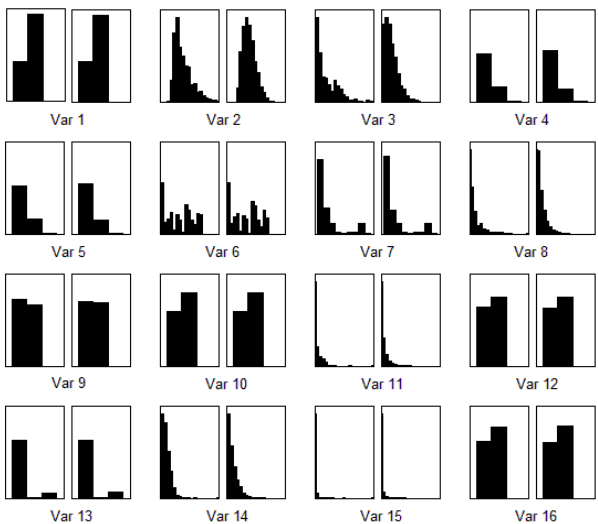

Fig. 12. Marginal distributions for original (left) and randomly generated (right) vectors for the Australian Credit dataset. Variables 2, 3, 8, 14 and 15 are continuous; all others are discrete.

*F. Clustering*

As a final application we consider the task of clustering. The clustering algorithm that we have developed [18] can loosely be thought of as a graph version of the mixture model approach [27], in which the data is modelled as a combination of components, with each component treated as a cluster. However, unlike Gaussian mixture models, which operate in a Euclidean space and use a likelihood function parameterized by the means and covariances of Gaussian components, which can be optimized using Expectation-Maximization [6], no explicit density model is used to represent clusters. Instead, clusters are represented as stationary distributions over the nodes in the graph. The problem is to discover the personalization vectors $\theta^k$ that give rise to these stationary distributions.

Recall that there are two types of nodes in our graphs: object nodes and attribute-value nodes. Since we are interested in clustering objects (and not attribute values), we will assume that the components of $\theta$ corresponding to attribute-value nodes will always be 0. Similarly, when we calculate the centrality vector, $c$, we will set the components corresponding to attribute-value nodes to 0, and normalize the remaining components to sum to 1.

Let $y_i^k$ represent the membership of object $i$ to cluster $k$. The membership values must satisfy the conditions

$$\sum_{k=1}^{K} y_i^k = 1$$
$$y_i^k \geq 0, \quad i = 1,...,N; \; k = 1,...,K$$

where $N$ is the number of object nodes, and $K$ is the number of clusters. The clustering algorithm is given below.

---
**ALGORITHM 2.** Clustering Algorithm
---
**Initialize** $\{\theta^k : k = 1, ..., K\}$ with random values
**Repeat**
  // Compute the first eigenvector $\mathbf{c}^k$ of the equation
  $\mathbf{c}^k = d\,\mathbf{W}\mathbf{c}^k + (1-d)\theta^k$, $(k = 1,...,K)$
  // Update cluster membership values
  $y_i^k = c_i^k \Big/ \sum_{j=1}^{K} c_i^j$, $(i = 1,...,N; k = 1,...,K)$
  // Update personalization vectors
  $\theta_i^k = \left(c_i^k y_i^k\right) \Big/ \sum_{j=1}^{N} \left(c_j^k y_j^k\right)$, $(i = 1,...,N; k = 1,...,K)$
**until convergence**

---

Step 1 calculates the stationary distribution for each cluster based on the personalization vector for that cluster. In Step 2, the membership values for each data point $i$ are determined by normalizing the centrality vector components to sum to 1 across all clusters. In Step 3, the personalization vectors are updated by scaling the centrality vector component with the membership values, and normalizing so that the personalization vector components for each cluster sum to 1. When the algorithm has converged, the stationary distribution vectors $\mathbf{c}^k$ will be equal to the personalization vectors $\theta^k$. If hard clustering is required, then a node can simply be assigned to the cluster for which its membership is greatest. The degree of cluster overlap can be controlled by varying the value of $d$ that appears in the eigenvector equation shown in Step 1. The smaller the value of $d$, the crisper (i.e., less overlap in) the clustering.

Results of applying the clustering method to the Australian Credit dataset are presented in Table III, together with results from three other clustering algorithms. Results for *k*-means, fuzzy *c*-means, non-negative matrix factorization are based on representing the mixed-variable data in the format described for matrix **B** in Section II. The clustering performance of our graph-based method measured using *V*-measure, Rand Index and *F*-measure is superior to that of *k*-means clustering, identical to that of Fuzzy *c*-means, and slightly inferior to that of non-negative matrix factorization using the approach in [28]. Additional results and discussion on the clustering algorithm can be found in [18].

TABLE III. CLUSTERING RESULTS

| Dataset | *V*-meas. | Rand | *F*-meas. |
|---|---|---|---|
| Graph-based Clustering | 0.294 | 0.690 | 0.692 |
| *k*-means | 0.156 | 0.604 | 0.619 |
| Fuzzy *c*-means | 0.294 | 0.690 | 0.692 |
| Non-neg. Mat. Fact. | 0.322 | 0.706 | 0.708 |

## VIII. CONCLUSION

This paper has presented a non-parametric method for estimating various probability distributions from datasets that contain a mix of continuous and discrete variables. The data are represented as a graph; an eigenvector equation is then constructed, based on the distribution to be estimated; the equation is then solved to find the stationary distribution, from which the required probabilities can then be obtained. These probability estimates can be used to perform a broad range of useful tasks including classification, regression, missing value imputation, outlier detection, random vector generation, and clustering.

The method requires one eigenvector equation to be solved for every value that is to be estimated. The eigenvector equation is solved using the power iteration method, whose rate of convergence depends heavily on the value of the parameter $d$ that appears in (4). In our experiments, we have used a value of 0.15, and found that convergence is achieved after approximately 50 iterations, which is between one and two orders of magnitude faster that for a value of 0.85 (a typical value used in PageRank), and this is despite there being no appreciable difference in the estimated distributions.

Like the $k$-nearest neighbour classification and $k$-means clustering algorithms, the method is memory-based; that is, the dataset must be stored in memory, and this may be a limiting factor on very large datasets. However, in such cases the method could itself be used to generate from the dataset a smaller set of random *prototype vectors* that have the same correlations and marginal distributions as the original dataset. This set of prototypes could then be used in place of the original dataset in subsequent applications.

The power and flexibility of the model arises out of its non-parametric nature, together with the graph-based representation of objects and their attributes. The non-parametric nature of the model means virtually no assumptions are made about the data. For example, there is no assumption that the marginal or conditional distributions follow any standard distribution (e.g., Gaussian) or that they are composed of any mixture of such distributions. This is in stark contrast to model-based approaches such as conditional Gaussian distributions [11], [12]. Another important feature of the model is that it does not require any explicit measure of distance or similarity between objects to be defined. Rather, any measure of distance or similarity is entirely implicit in—and inseparable from—the combination of graph-based representation and eigencentrality measure used to calculate the stationary distribution. This is particularly attractive in the case of mixed-variable datasets such as the Australian Credit dataset, in which a good measure of similarity or distance may be difficult to define.

While we have presented the method as a method for dealing with mixed-variable data, it can also be applied to continuous-only and discrete-only data. It could also be easily applied to text data. For example, documents (or paragraphs or sentences) could be treated as objects, words could be treated as attributes, and the weights may be tf-idf scores.

## REFERENCES

[1] C. Bishop, Pattern Recognition and Machine Learning. Springer, 2006.

[2] C.K. Enders, Applied missing data analysis. Guilford Press, 2010.

[3] A. Zimek, E. Schubert and H.P. Kriegel, "A survey on unsupervised outlier detection in high-dimensional numerical data," Statistical Analysis and Data Mining, vol. 5, no. 5, pp. 363–387, 2012.

[4] M.E. Johnson, 1987. Multivariate Statistical Simulation. John Wiley & Sons, 1987.

[5] A.M. Law and W.D. Kelton. Simulation Modeling and Analysis. McGraw-Hill, 2000.

[6] A.P. Dempster, N.M. Laird and D.B. Rubin, "Maximum likelihood from incomplete data via the EM algorithm," Journal of the Royal Statistical Society. Series B (Methodological), vol. 39, no. 1, pp. 1-38, 1977.

[7] B.S. Everitt and G. Dunn, Applied Multivariate Data Analysis (Vol. 2). Arnold, 2001.

[8] J.R. Quinlan, "Simplifying decision trees," International Journal of Man-Machine Studies, vol. 27, no. 3, pp. 221-234, 1987.

[9] A.R. de Leon and K.C. Chough, "Analysis of mixed data: methods and applications: an overview," Analysis of Mixed Data: Methods and Applications, A.R. de Leon and K.C. Chough, eds., CRC Press, pp, 1-11, 2013.

[10] D. Edwards, Introduction to Graphical Modelling. Springer Science & Business Media, 2012.

[11] S.L. Lauritzen and N. Wermuth, "Graphical models for associations between variables, some of which are qualitative and some quantitative," The Annals of Statistics, pp.31-57, 1989.

[12] S. Lauritzen, Graphical Models., Oxford University Press, 1996.

[13] D. Edwards, "Hierarchical interaction models," Journal of the Royal Statistical Society, Series B, vol. 52, pp. 3-20, 1990.

[14] J. Lee and T. Hastie, "Structure learning of mixed graphical models," Artificial Intelligence and Statistics, pp. 388-396, 2013.

[15] B. Fellinghauer, P. Buhlmann, M. Ryffel, M. Von Rhein and J.D. Reinhardt, "stable graphical model estimation with random forests for discrete, continuous and mixed variables," Computational Statistics & Data Analysis, vol. 64, pp. 132-152, 2013.

[16] J. Cheng, T. Li, E. Levina and J. Zhu, "High-dimensional mixed graphical models. arXiv preprint arXiv:1304.2810, 2013.

[17] A. Skabar, "Random Vector Generation from Mixed-Attribute Datasets using Random Walk," *Proc. Winter Simulation Conference (WSC '16)*, pp. 988-997, 2016.

[18] A. Skabar, "Clustering Mixed-Attribute Data using Random Walk," *Proc. International Conference on Computational Science (ICCS 2017)*, pp. 1096-1107, 2017.

[19] M.E.J. Newman, Networks: an Introduction. Oxford University Press, 2010.

[20] O. Peron, (1907) "Zur theorie der matrices", Mathematische Annalen, vol. 64, no. 2, pp. 248–263, 1907.

[21] G. Frobenius, "Ueber matrizen aus nicht negativen elementen", Sitzungsberichte der Königlich Preussischen Akademie der Wissenschaften, pp. 456–477, 1912.

[22] L. Katz, "A new status index derived from sociometric index. Psychometrika, vol. 18, pp. 39-43, 1953.

[23] S. Brin and L. Page, "The anatomy of a large-scale hypertextual web search engine," Computer Networks and ISDN Systems, vol. 30, pp. 107-117, 1998.

[24] G. Jeh and J. Widom, "Scaling personalized web search," Proceedings of the 12th International Conference on World Wide Web, pp. 271-279, 2003.

[25] J.Y. Pan, H.J. Yang, C. Faloutsos and P. Duygulu, "Automatic multimedia cross-modal correlation discovery," Proceedings of the tenth ACM SIGKDD international conference on Knowledge discovery and data mining, pp. 653-658, 2004.

[26] S. Geman and D. Geman, "Stochastic relaxation, Gibbs distributions, and the Bayesian restoration of images," IEEE Trans. Pattern Analysis and Machine Intelligence, vol. 6, pp. 721-741, 1984.

[27] D. Titterington, A. Smith and U. Makov, Statistical Analysis of Finite Mixture Distributions. Wiley, 1995.

[28] D.D. Lee & H.S. Seung, "Learning the parts of objects by non-negative matrix factorization," Nature, 401(6755): 788-791, 1999.